\pdfoutput=1

\documentclass[11pt]{article}

\usepackage[final]{acl}

\usepackage{times}
\usepackage{latexsym}

\usepackage[T1]{fontenc}

\usepackage[utf8]{inputenc}

\usepackage{microtype}

\usepackage{inconsolata}

\usepackage{graphicx}
\usepackage{amsmath}
\usepackage{booktabs}
\usepackage{multirow}
\usepackage{amsfonts}

\title{RAR$^2$: Retrieval-Augmented Medical Reasoning \\ via Thought-Driven Retrieval}

\author{Kaishuai Xu$^{1}$, Wenjun Hou$^{1,2\ast}$, Yi Cheng$^{1\ast}$, Wenjie Li$^{1}$\\
$^1$Department of Computing, The Hong Kong Polytechnic University, Hong Kong \\
$^2$Research Institute of Trustworthy Autonomous Systems and \\Department of Computer Science and Engineering, \\
Southern University of Science and Technology, Shenzhen, China \\
\texttt{\{kaishuaii.xu, alyssa.cheng\}@connect.polyu.hk,} \\
\texttt{houwenjun060@gmail.com, cswjli@comp.polyu.edu.hk}
}

\begin{document}
\maketitle

\begin{abstract}

Large Language Models (LLMs) have shown promising performance on diverse medical benchmarks, highlighting their potential in supporting real-world clinical tasks. Retrieval-Augmented Generation (RAG) has emerged as a key approach for mitigating knowledge gaps and hallucinations by incorporating external medical information. However, RAG still struggles with complex medical questions that require intensive reasoning, as surface-level input often fails to reflect the true knowledge needs of the task. Existing methods typically focus on refining queries without explicitly modeling the reasoning process, limiting their ability to retrieve and integrate clinically relevant knowledge. In this work, we propose RAR$^2$, a joint learning framework that improves both Reasoning-Augmented Retrieval and Retrieval-Augmented Reasoning. RAR$^2$ constructs a thought process to uncover implicit knowledge requirements and uses it to guide retrieval and answer generation. We build a training dataset of mixed preference pairs and apply Direct Preference Optimization (DPO) to train the model. Moreover, we design two test-time scaling strategies to explore the boundaries of our framework. Experiments demonstrate the effectiveness of RAR$^2$ across several biomedical question answering datasets, outperforming RAG baselines with or without fine-tuning. 

\end{abstract}
\section{Introduction}

\begin{figure}[th!]
	\centering
	\includegraphics[width=1.0\linewidth]{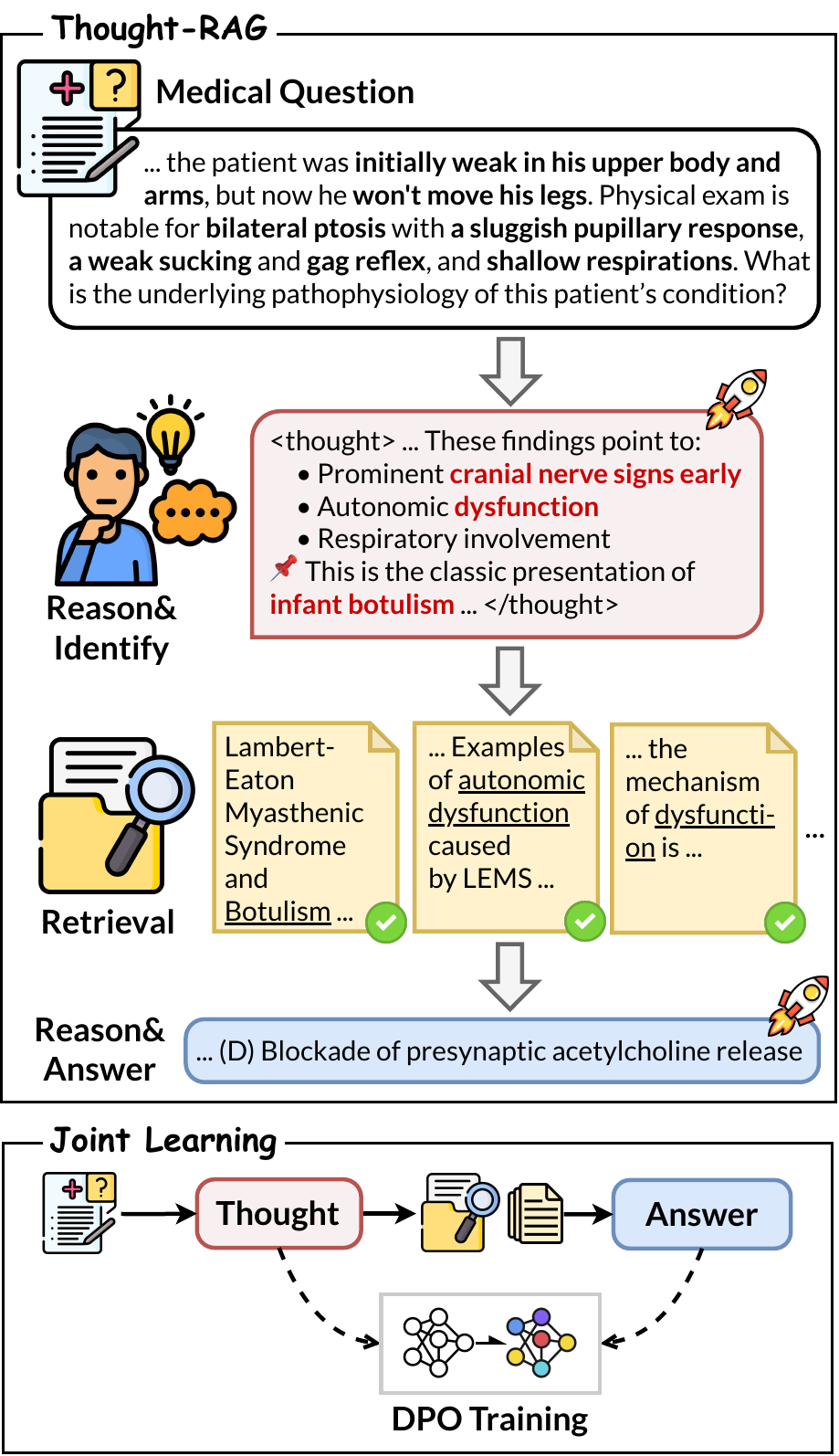}
	\caption{Thought-RAG for medical reasoning. Our method jointly optimizes the generation of thought processes and the subsequent retrieval-augmented answer generation.}
	\label{example}
\end{figure}

The capabilities of Large Language Models (LLMs) in medicine have long attracted substantial research attention \cite{med-palm, med-palm2, baichuan-m1}. LLMs such as GPT-4o and Baichuan-M1 have demonstrated strong performance across diverse medical benchmarks, highlighting their potential to support real-world clinical tasks \cite{challenging-medqa, baichuan-m1, emulation}. More recently, given the vast scope and high-stakes nature of the medical domain, Retrieval-Augmented Generation (RAG) has emerged as a key approach for mitigating knowledge gaps and hallucinations by leveraging external medical knowledge \cite{med-r2, med_textbooks, medical-graph-rag, rag-healthcare}. 

While RAG performs well in a range of medical tasks, it still struggles with complex medical questions requiring intensive reasoning \cite{bright}. This difficulty arises because the surface-level information in these questions does not reflect their actual knowledge requirements. For example, in Figure \ref{example}, although the question clearly describes the patient’s symptoms, such as ``weak in upper body and arms'' and ``won’t move legs'', the knowledge necessary to generate the correct answer lies in clinical concepts like ``autonomic dysfunction'' and ``infant botulism''. These concepts must be inferred through detailed analytical reasoning. Existing RAG methods for complex medical reasoning primarily focus on direct query refinement, but fail to construct and optimize a reasoning process that comprehensively uncovers the underlying knowledge requirements \cite{med-r2, rgar, serts}. 

Additionally, few studies have explored enhancing the medical reasoning capabilities of LLMs within the RAG framework \cite{self-biorag}. Reasoning with external knowledge is more challenging, as the retrieved content inevitably includes noise information. LLMs should learn to integrate relevant knowledge into the reasoning process while avoiding interference from irrelevant information. Therefore, optimizing retrieval-augmented reasoning is essential for improving both the accuracy and robustness of generation outcomes.

In this work, we construct a thought process to reason through medical questions and identify their implicit knowledge requirements. This thought process is directly used to retrieve relevant information, which is subsequently incorporated into reasoning to derive the final answer. To optimize both thought and answer generation, we propose a joint learning framework, \textbf{RAR$^2$}, that simultaneously improves \textbf{R}easoning-\textbf{A}ugmented \textbf{R}etrieval and \textbf{R}etrieval-\textbf{A}ugmented \textbf{R}easoning. Specifically, we first construct a training dataset consisting of mixed preference pairs. One type is thought pairs, in which a sampled thought process is annotated based on the outcome of subsequent retrieval-augmented generation. The other type is answer pairs, in which a sampled answer is annotated according to its correctness. We then apply Direct Preference Optimization (DPO) to fine-tune the LLM \cite{dpo}. These preference pairs enable supervised preference learning, allowing the model to identify relevant knowledge and reason effectively with external information. Extensive experiments across several biomedical question answering datasets demonstrate RAR$^2$'s superiority over existing RAG baselines. Our further test-time scaling analysis validates the scalability of RAR$^2$. 

In summary, our contributions are as follows:
\begin{itemize}
	\item We propose a joint learning framework, RAR$^2$, that simultaneously improves reasoning-augmented retrieval and retrieval-augmented reasoning. Additionally, we design two test-time scaling strategies to explore the boundaries of our framework. 
	\item We construct a mixed preference dataset to train the LLM to identify implicit knowledge needs and reason effectively with external information.
	\item Experimental results demonstrate the effectiveness of RAR$^2$ in six biomedical question answering datasets. Our framework outperforms the baseline under tuning-free and fine-tuned settings.
\end{itemize}

\section{Preliminary}

\subsection{Problem Formulation}
In this work, we focus on LLM-based medical reasoning for medical question answering. Given a question $q$ and a large medical corpus $\mathcal{D} = \{d_1, d_2, \dots, d_n\}$, a set of top-relevant documents $\mathcal{D}_k$ is retrieved for $q$, where $n$ is the total number of documents in the corpus and $k$ is the number of retrieved ones. Then, the question and the retrieved documents are used as context for an LLM $\mathcal{M}$ to generate an answer $y$ by step-by-step reasoning. Our goal is to jointly optimize document retrieval and retrieval-augmented medical reasoning. 

\subsection{Medical Corpus}
A medical corpus is crucial for retrieval-augmented medical reasoning, as it provides the external knowledge necessary for LLMs to reason effectively about a given question. In this work, we adopt MedCorp, a large-scale medical corpus proposed by \citet{medrag}. MedCorp integrates four major sources: PubMed\footnote{\url{https://pubmed.ncbi.nlm.nih.gov/}}, StatPearls\footnote{\url{https://www.statpearls.com/}}, medical textbooks \cite{med_textbooks}, and Wikipedia. It comprises 30.4M documents, including clinical guidelines, peer-reviewed research articles, and medical encyclopedic content. To process the documents, we reuse the original chunking strategy provided with the corpus and denote each document chunk as $d_i \in \mathcal{D}$, where $\mathcal{D}$ is the set of all chunks.

\section{Reasoning Before Retrieval}

\begin{figure}[t!]
	\centering
	\includegraphics[width=1\linewidth]{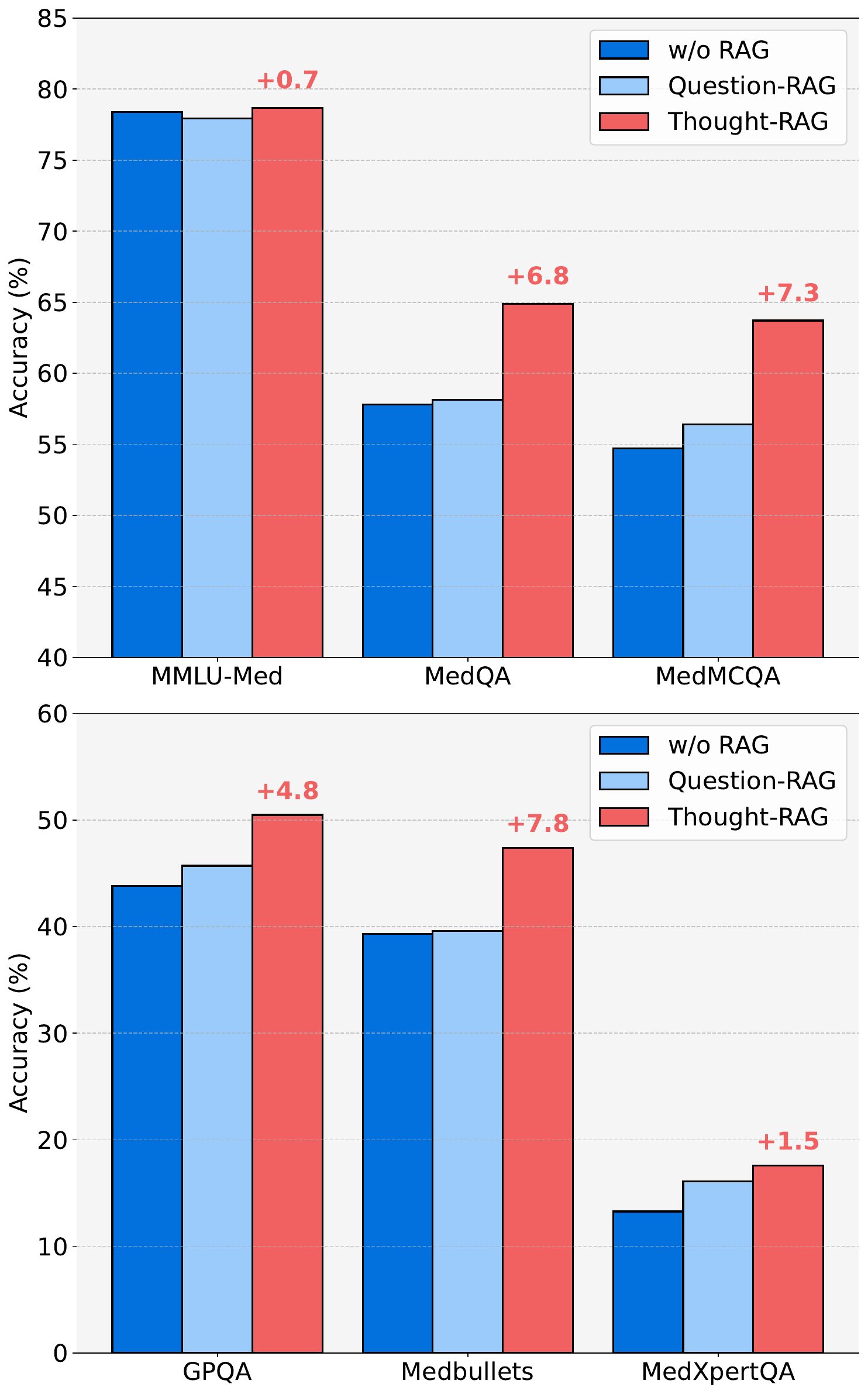}
	\caption{The performance comparison between Question-RAG and Thought-RAG on several medical question answering datasets.}
	\label{difference_rag}
\end{figure}

We begin with a preliminary study on the impact of reasoning prior to retrieval. Rather than fine-tuning a model to generate or refine a query, we instead sample a thought process that could guide the model toward a final answer and use it directly as a query to retrieve relevant documents. The thought process is sampled in a zero-shot manner by prompting the LLM with: ``\textit{Please reason step by step to identify the relevant knowledge that may be involved.}'' For retrieval, we use the BM25 algorithm \cite{bm25} to retrieve the top-k documents from the corpus. These documents are then provided as context to the LLM for answer generation. We evaluate this approach—referred to as Thought-RAG—against standard question-based retrieval (Question-RAG) using the Qwen2.5-7B-Instruct model \cite{qwen2.5} with $k=32$. 

Figure \ref{difference_rag} shows the performance comparison on several medical question answering datasets. We can observe that Thought-RAG consistently outperforms Question-RAG, with an average improvement of 4.82\% across all datasets. Especially in challenging datasets like Medbullets and GPQA, the gains are more significant. Furthermore, for most datasets, the accuracy remains unchanged or even declines using Question-RAG. A similar phenomenon can be found in the MIRAGE benchmark \cite{medrag}. These observations suggest that reasoning-intensive medical questions often do not explicitly state the specific information needs required for successful retrieval. In contrast, a thought process generated through prior reasoning more accurately captures the underlying knowledge requirements, thereby improving overall RAG performance. Reasoning first and then using the resulting thought process for retrieval is a straightforward yet powerful approach to collect relevant knowledge, but it has received limited attention within the RAG community \cite{rationale-guided-rag}. Therefore, further optimizing the generation of the thought process holds promise for enhancing retrieval quality and improving the RAG performance. 

\section{Method}

\begin{figure*}[th!]
	\centering
	\includegraphics[width=1\linewidth]{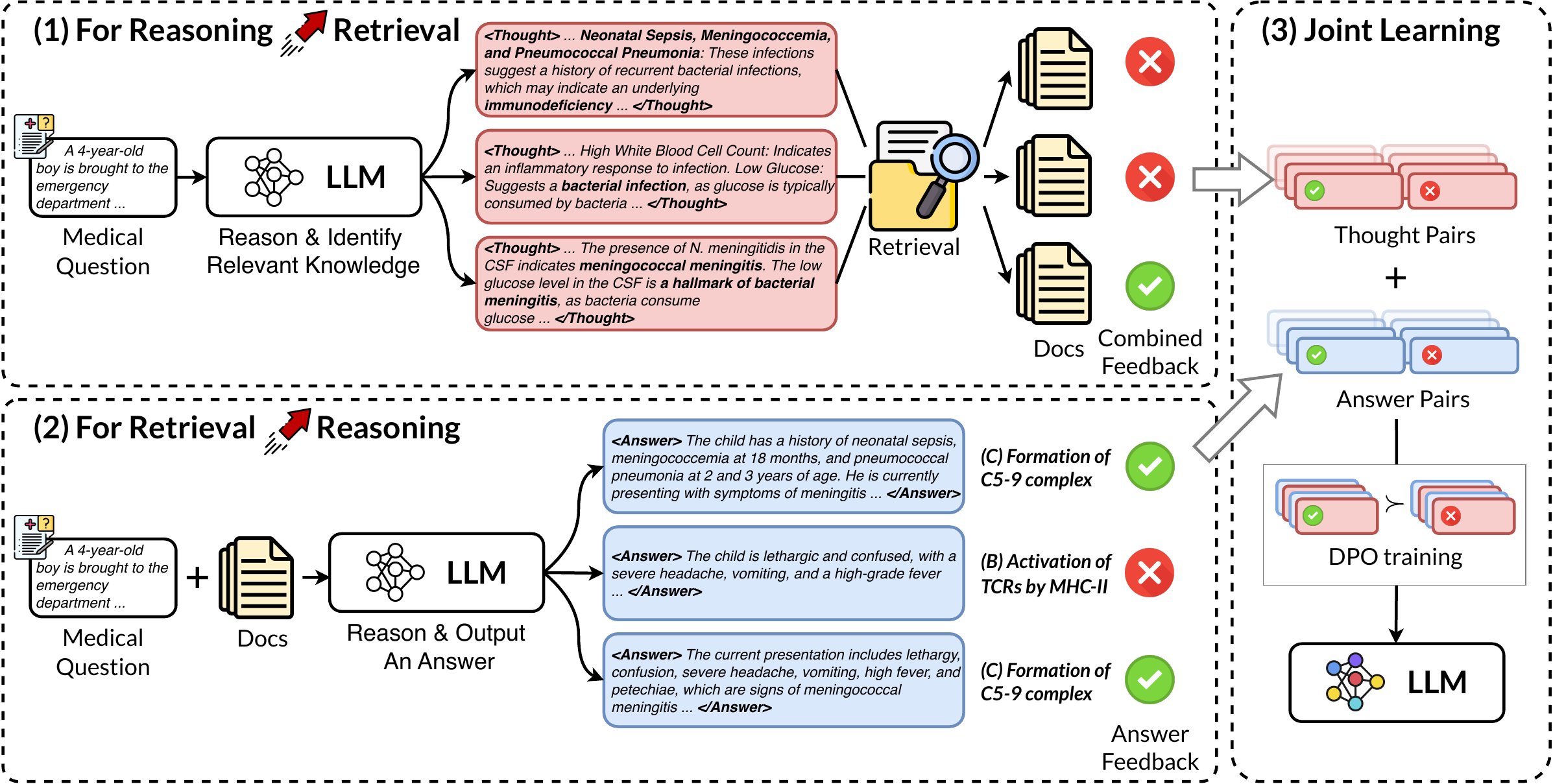}
	\caption{Construction of mixed preference pairs. (1) We sample several thought processes and append each to the question to form a query for retrieval. Each thought is then annotated based on whether it leads to a correct answer in subsequent RAG. (2) We sample several answers and annotate each one according to its correctness.}
	\label{construction}
\end{figure*}

In this section, we introduce the RAR$^2$ framework, which constructs a mixed preference dataset consisting of thought and answer pairs, and applies DPO to jointly enhance both reasoning-augmented retrieval and retrieval-augmented reasoning for LLMs. As shown in Figure \ref{construction}, we first sample a set of thought processes and answers. The thought processes are generated based on the medical question, while the answers are generated using the question along with the corresponding retrieved documents. All samples are annotated based on the correctness of the answer (§\ref{mixed-pairs}). We collect samples to form a mixed preference dataset and apply DPO to jointly optimize two types of reasoning processes (§\ref{joint-learning}). Besides, we design two test-time scaling strategies to examine the boundaries of RAR$^2$ (§\ref{test-time-scaling}). 

\subsection{Construction of Mixed Preference Pairs}\label{mixed-pairs}
Our goal is to formulate and optimize two types of reasoning processes: one for retrieval and the other for retrieval-augmented generation. Previous work often overlooks the reasoning process required to investigate the underlying knowledge requirements, and few studies focus on jointly optimizing both retrieval and generation. To address this, we design distinct preference pairs tailored to each type of reasoning process. 

\paragraph{Thought Pairs.} We use an instruction-tuned LLM as the base model to sample thought processes. Given a medical question $q$, we first prompt the model with ``\textit{Please reason step by step to identify the relevant knowledge that may be involved.}'' to sample several thought processes: 
\begin{equation}
	y^t \sim \mathcal{M}(q, \text{Prompt}_t).
\end{equation}
We then annotate them as chosen and rejected based on two criteria. The first criterion appends the prompt ``\textit{The answer is: }'' to the end of each thought process and lets the LLM complete the answer. The correctness of the result is denoted as $g^{\text{direct}} \in \{0, 1\}$. The second criterion uses the thought process as a query to retrieve the top-$k$ documents, which are then used to generate an answer. The correctness of this result is denoted as $g^{\text{rag}} \in \{0, 1\}$. To ensure deterministic outputs, we set the temperature to 0 during annotation. The thought samples are labeled as chosen only if both $g^{\text{direct}} = 1$ and $g^{\text{rag}} = 1$. Such an annotation ensures that the thought process can lead to a correct answer. We collect one chosen and one rejected thought process to form a preference pair, denoted as $(y^{t+}, y^{t-})$. 

\paragraph{Answer Pairs.} We use the same base model to sample answer candidates. Given a medical question $q$ and a thought process $y^t$, we first retrieve top-$k$ documents $\mathcal{D}_k$ from the corpus. Then, we sample a group of answers based on the question and each retrieved document set:
\begin{equation}
	y^a \sim \mathcal{M}(q, \mathcal{D}_k, \text{Prompt}_a),
\end{equation}
where $\text{Prompt}_a$ is ``\textit{Please reason step by step and choose one option from the above}''. The sample is labeled as chosen if its answer is correct. We collect one chosen and one rejected answer to form a preference pair, denoted as $(y^{a+}, y^{a-})$. 

\subsection{Joint Learning}\label{joint-learning}
After collecting the thought and answer pairs, we construct a mixed preference dataset $\mathcal{Y}=\{(y^{t+}, y^{t-})\} \cup \{(y^{a+}, y^{a-})\}$. We then apply DPO to jointly optimize both types of reasoning processes. Joint training is adopted because the two processes are complementary and can benefit from each other: reasoning to identify relevant knowledge facilitates answer generation, while answer generation, in turn, helps refine the analytical quality of the thought process. Previous studies fail to optimize these reasoning processes or consider their complementary relationship. 

For DPO training, we thoroughly shuffle all preference pairs and employ the DPO loss as follows:
\begin{align}
	&\mathcal{L} = -\mathbb{E}_{(q,y^+,y^-) \sim \mathcal{Y}} \big[ \log \sigma \big( \nonumber \\
	&\beta \log \frac{\pi_\theta(y^+ | q)}{\pi_{\mathcal{M}}(y^+ | q)} - \beta \log \frac{\pi_\theta(y^- | q)}{\pi_{\mathcal{M}}(y^- | q)}
	\big) \big],
\end{align}
where $\sigma$ is the sigmoid function, $\pi_{\theta}$ is the policy model, $\pi_{\mathcal{M}}$ is the reference model, and $\beta$ is the hyperparameter that regulates the extent of deviation from the reference model. 

\subsection{Test-Time Scaling}\label{test-time-scaling}

\begin{figure}[t!]
	\centering
	\includegraphics[width=1\linewidth]{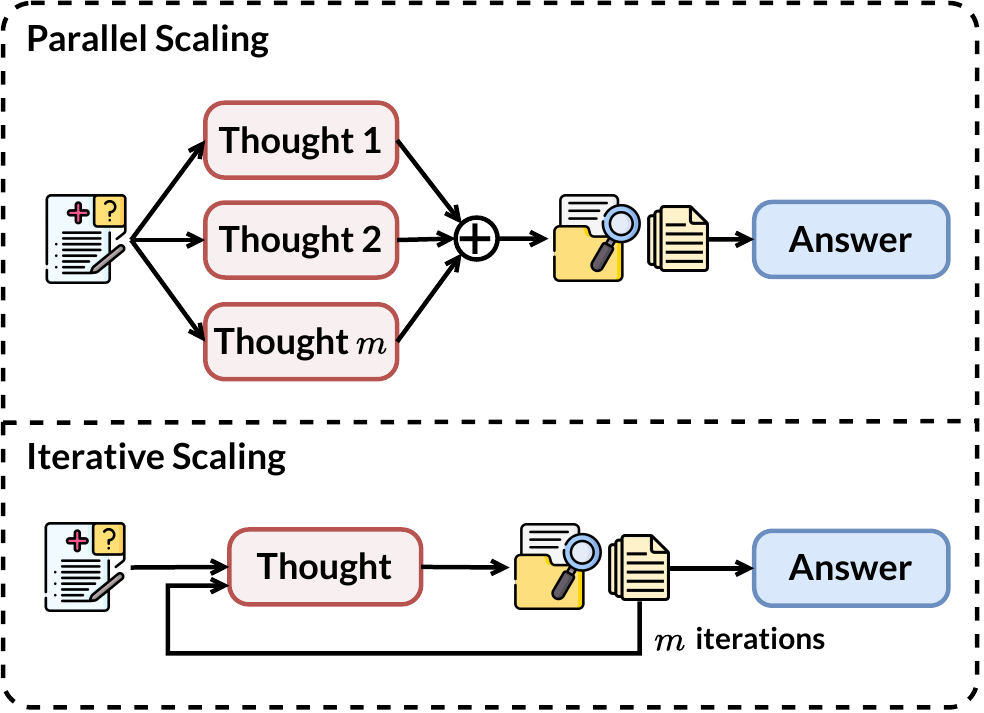}
	\caption{Frameworks of test-time scaling. Parallel scaling concatenates multiple thought processes to retrieve documents, while iterative scaling uses the retrieved documents in sequence to enhance thought generation.}
	\label{scaling}
\end{figure}

To investigate the test-time scaling effect of RAR$^2$, we design two scaling strategies, as illustrated in Figure \ref{scaling}. Previous studies have paid little attention to the continual improvement of medical reasoning within the RAG framework.

The first strategy (\textbf{Paralleling Scaling}) generates $m$ thought processes in parallel and concatenates them to form a single extended thought process. This combined thought is then used to retrieve documents, based on which an answer is generated. The second strategy (\textbf{Iterative Scaling}) generates a thought process and retrieves documents, which are then appended to the prompt for the next round of thought generation. This iterative process is repeated $m$ times, after which an answer is generated. We design these two strategies to explore how RAR$^2$ scales with the number of generated thought processes. The first strategy merges more useful information to retrieve documents, while the second exploits the retrieved documents to generate more accurate thought processes. 

\section{Experiments}

\subsection{Evaluation Datasets}
We evaluate our method on six biomedical question answering datasets: MedQA, MedMCQA \cite{medmcqa}, MMLU-Med (MMLU.) \cite{mmlu}, Medbullets \cite{challenging-medqa}, GPQA \cite{gpqa}, and MedXpertQA \cite{medxpertqa}. Among these, MedQA is the in-domain dataset, as our training data is derived from it. The remaining datasets are considered out-of-domain. The number of answer options across datasets ranges from 4 to 10. Notably, Medbullets, GPQA, and MedXpertQA are more recent and consist of graduate- or expert-level problems. Overall, the diversity of these datasets enables a robust evaluation of the model’s medical reasoning capabilities.

\subsection{Baseline Methods}
We compare our framework with six strong baselines in medical reasoning. Qwen2.5-7B-Instruct \cite{qwen2.5} is used as our base model. Two recent models enhanced for medical reasoning that do not use RAG are included: m1-7B-23K \cite{m1} and HuatuoGPT-o1 \cite{huatuogpt-o1}. For a fair comparison, we use HuatuoGPT-o1-7B with the same model size. Additionally, we include three tuning-free RAG methods that have been applied to medical tasks: MedRAG \cite{medrag}, i-MedRAG \cite{i-medrag}, and Med-R$^2$ \cite{med-r2}. Med-R$^2$ is developed through a few-shot generation strategy. Finally, we include four RAG methods that involve fine-tuning on specific components: Self-BioRAG \cite{self-biorag}, SimRAG-8B \cite{simrag}, RAG$^2$-7B \cite{med-r2}, and SPO \cite{omni-rag}. Baselines with publicly released models are reimplemented for comparison. 

\subsection{Implementation Details}

\begin{table*}[t!]
    \centering
    \resizebox{1.0\textwidth}{!}{
    \begin{tabular}{@{}lccccccc@{}}
    \toprule
    Methods              & MedQA† & MedMCQA & MMLU. & GPQA  & Medbullets & MedXpertQA & Avg. \\ \midrule
    \multicolumn{8}{c}{\textit{No RAG}}                                                \\ \midrule
    Qwen2.5-7B           & 57.82 & 54.70   & 78.39    & 43.81 & 39.29      & 13.27      & 47.20   \\ 
    m1-7B                & 64.34 & 59.34   & 78.02    & 36.19 & 48.38      & 16.29      & 50.43   \\ 
    HuatuoGPT-o1         & 68.81 & 64.95   & 79.87    & 45.71 & 50.65      & 14.98      & 54.17   \\ \midrule
    \multicolumn{8}{c}{\textit{Tuning-Free RAG}}                                                   \\ \midrule
    MedRAG               & 54.20 & 52.35   & 75.81    & 42.86 & 39.29      & 14.20      & 46.45   \\
    i-MedRAG             & 62.84 & \underline{55.18}  & \textbf{79.87}    & \underline{51.43} & \underline{44.81}      & \underline{16.37}       & \textbf{51.75}       \\
    Med-R$^2$            & \textbf{81.06} & 49.27   & 72.39    & -     & -          & -          & -       \\
    RAR$^2$ (w/o train)  & \underline{64.89} & \textbf{63.71}   & \underline{78.67}    & \underline{50.48} & \textbf{47.40}      & \textbf{17.59}      & \textbf{54.27}   \\ \bottomrule
    \end{tabular}
    }
    \caption{Comparisons of RAR$^2$ with other medical large language models and tuning-free RAG methods. † denotes the in-domain dataset.}
    \label{main_results1}
    \end{table*}

We utilize the MedQA training set \cite{medqa} to implement our method, which comprises 10K medical problems derived from professional medical board exams. Each problem requires selecting the correct answer from four options. Our base model for sampling preference pairs is Qwen2.5-7B-Instruct \cite{qwen2.5}. We set the number of thought sampling attempts to 15 and the number of answer sampling attempts to 5. The temperature of sampling is set to 0.2, and the top-$p$ is set to 0.9. For each question, we retrieve the top 32 document chunks using the BM25 algorithm \cite{bm25} with parameters $k_1=1.2$ and $b=0.75$ as context for sampling solutions. A total of 12K preference pairs are collected after filtering and selection, where 4K are thought pairs and 8K are answer pairs. The model is then trained for 4 epochs with a global batch size of 64 and a learning rate of 1e-6, while the parameter $\beta$ for the DPO loss is set to 0.2. All experiments are conducted on eight A100 GPUs, and we employ DeepSpeed ZeRO3 to optimize memory usage.

\subsection{Main Results}

\begin{table}[t!]
    \centering
    \resizebox{1.0\linewidth}{!}{
    \begin{tabular}{@{}lccc@{}}
    \toprule
    Methods      & MedQA† & MedMCQA & MMLU.  \\ \midrule
    Self-BioRAG  & 43.60 & 42.15   & 53.92     \\ 
    SimRAG-8B    & 62.92 & 67.51   & 75.57     \\ 
    RAG$^2$-7B   & 75.64 & 63.04   & 78.67     \\ 
    SPO          & \textbf{76.98} & \textbf{71.08}   & \underline{85.49}     \\
    RAR$^2$ (w/ train)   & \underline{76.43} & \underline{65.69}   & \textbf{86.32}   \\ \bottomrule
    \end{tabular}
    }
    \caption{Comparisons of RAR$^2$ with other RAG methods that involve fine-tuned components. † denotes the in-domain dataset.}
    \label{main_results2}
    \end{table}

We report the main results on six biomedical datasets shown in Table \ref{main_results1} and Table \ref{main_results2}. We calculate the average performance across several datasets as \textbf{Avg.} in the final column. We present the results of RAR$^2$ under two settings: \textbf{w/o train} and \textbf{w/ train}. The former uses the base model for inference, while the latter uses the DPO-tuned model.

Table \ref{main_results1} presents the results in comparison with medically enhanced LLMs and tuning-free RAG methods. Our framework significantly improves the medical reasoning capabilities of the LLM over its backbone model, Qwen2.5-7B. It outperforms the backbone on all six datasets, achieving an average accuracy gain of 7.07\%. Additionally, RAR$^2$ performs competitively with state-of-the-art medical language models: it surpasses m1-7B and achieves comparable average accuracy to HuatuoGPT-o1. Notably, on challenging MedXpertQA, RAR$^2$ achieves accuracy gains of 1.3\% and 2.61\% over m1-7B and HuatuoGPT-o1, respectively. Compared to other tuning-free RAG methods, RAR$^2$ also achieves higher average accuracy. Specifically, it outperforms i-MedRAG, which uses iterative query refinement, and Med-R$^2$, which employs a more complex retrieval pipeline. 

Table \ref{main_results2} presents the results in comparison with other RAG methods that are applied to medical tasks and involve fine-tuned components. Our RAR$^2$ outperforms most baseline methods. Notably, on MMLU-Med, it achieves the highest accuracy among all models. SPO performs well on MedQA and MedMCQA, which may be attributed to its use of larger and open-source medical knowledge sources and additional test sample selection. Overall, the results on both in-domain and out-of-domain datasets demonstrate that our framework can help general LLMs consistently improve their medical reasoning abilities for both retrieval and answer generation. 

\subsection{Ablation Studies}

\begin{table*}[t!]
    \centering
    \resizebox{1.0\textwidth}{!}{
    \begin{tabular}{@{}lccccccc@{}}
    \toprule
    Methods        & MedQA & MedMCQA & MMLU.    & GPQA  & Medbullets & MedXpertQA & Avg. \\ \midrule
    RAR$^2$        & 76.43 & 65.69   & 86.32    & 56.19 & 57.14      & 20.98      & 60.46   \\ \midrule
    - w/o training & 64.89 & 63.71   & 78.67    & 50.48 & 47.40      & 17.59      & 54.27   \\ 
    - w/o answer   & 74.39 & 63.88   & 83.75    & 53.33 & 53.57      & 18.90      & 57.97   \\ 
    - w/o thought  & 74.86 & 64.79   & 84.85    & 54.29 & 55.19      & 19.51      & 58.92   \\ \bottomrule
    \end{tabular}
    }
    \caption{Ablation study on several medical question answering datasets.}
    \label{ablation_results}
    \end{table*}

We demonstrate the effectiveness of RAR$^2$ under different training settings, as detailed below: (1) \textbf{w/o training}, which adopts the thought-based RAG framework without any fine-tuning; (2) \textbf{w/o answer}, which removes answer pairs and optimizes only the generation of thought processes; and (3) \textbf{w/o thought}, which removes thought pairs and optimizes only the generation of answers. The results are shown in Table \ref{ablation_results}.

As shown in the table, RAR$^2$ achieves the best performance across all three selected datasets, demonstrating the effectiveness of our proposed method. Compared to the setting without training, both \textit{w/o answer} and \textit{w/o thought} yield significant performance improvements, indicating that optimizing the generation of either thought processes or answers is crucial for RAR$^2$’s effectiveness. More importantly, optimizing both types of preference pairs yields complementary gains, enhancing the performance of both reasoning-augmented retrieval (i.e., thought generation) and retrieval-augmented reasoning (i.e., answer generation).

\begin{figure}[t!]
	\centering
	\includegraphics[width=1\linewidth]{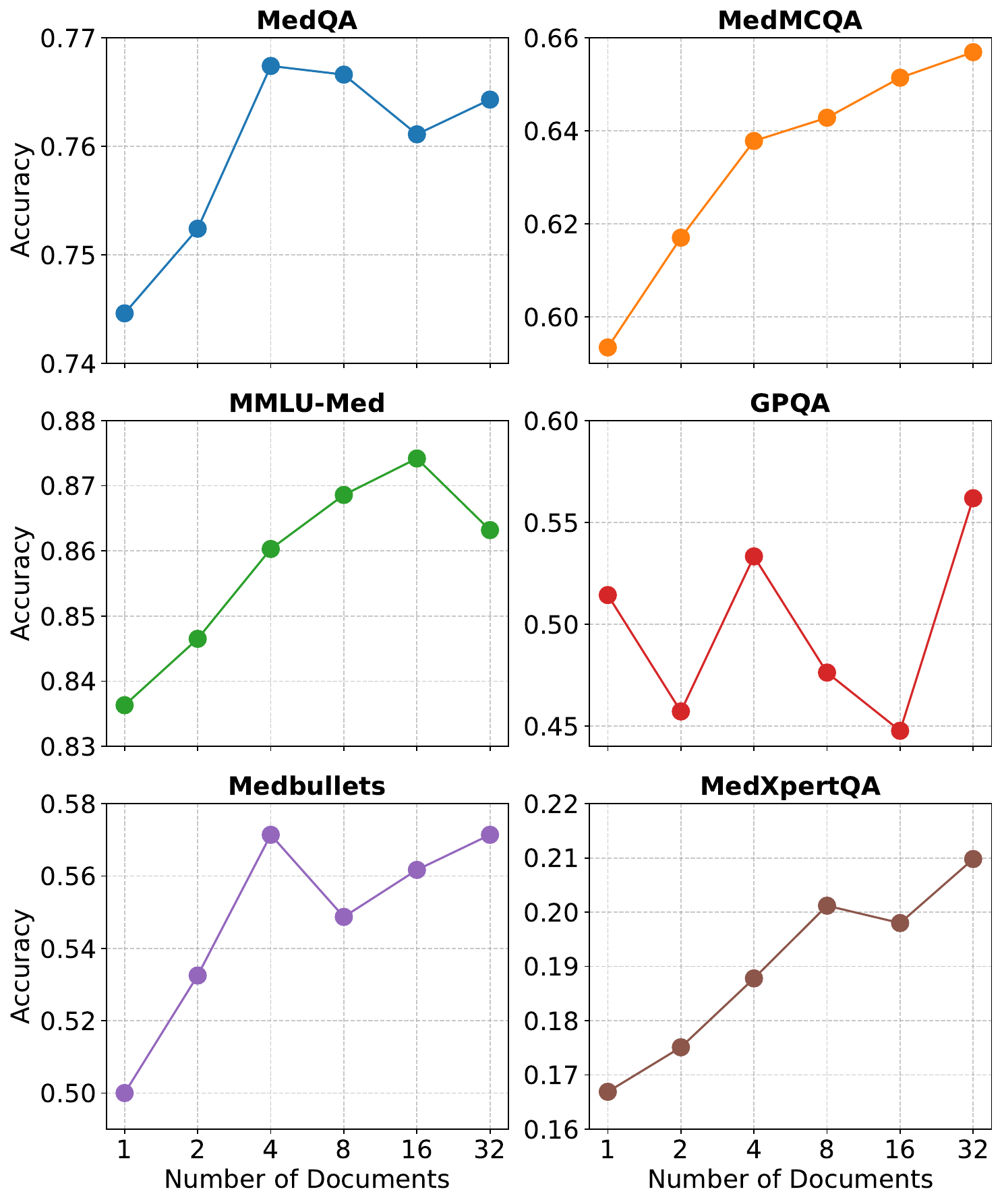}
	\caption{The performance comparison with a different number of retrieved documents.}
	\label{top_k}
\end{figure}

We also investigate the impact of the number of retrieved documents on the performance of RAR$^2$. As shown in Figure \ref{top_k}, RAR$^2$ achieves its best overall performance when retrieving 32 documents, although the optimal number varies slightly across different datasets. 

\subsection{Impact of Test-Time Scaling}

\begin{figure*}[t!]
	\centering
	\includegraphics[width=1.0\linewidth]{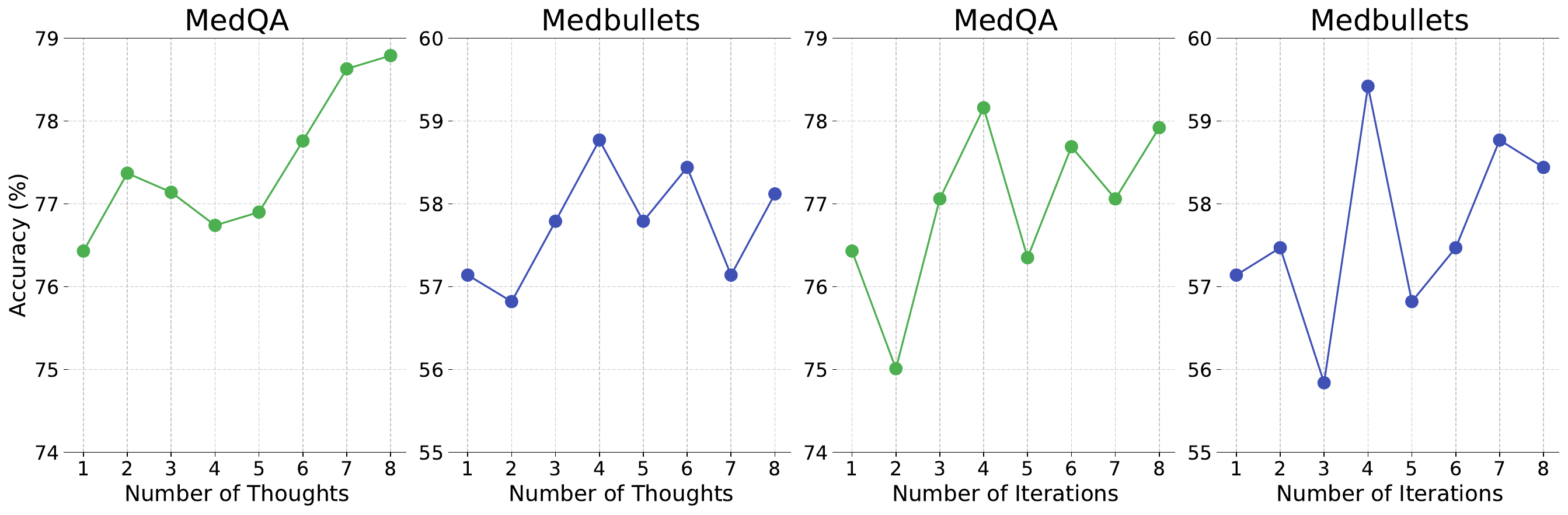}
	\caption{Accuracies with different scaling strategies on the MedQA and Medbullets datasets. The left two present Parallel Scaling, and the right two present Iterative Scaling.}
	\label{scaling_results}
\end{figure*}

We investigate the impact of test-time scaling on RAR$^2$, as shown in Figure \ref{scaling_results}. The total number of generated thought processes is scaled from 1 to 8. For parallel scaling, the sampling temperature is set to 1.0, and the top-$p$ value is also set to 1.0. For iterative scaling, the number of thoughts depends on the number of iterations. 

Figure \ref{scaling_results} shows the results with increasing iterations on the MedQA and Medbullets datasets, where one is the in-domain dataset and the other is a challenging out-of-domain dataset. For Parallel Scaling, accuracy shows an upward trend as the number of thought processes increases. The improvement is particularly stable on MedQA. A similar effect is observed in Iterative Scaling, where accuracy generally increases with the number of iterations, achieving a maximum improvement of approximately 2\% on both datasets.

\section{Related Work}

\subsection{Medical Reasoning}

Medical reasoning is a critical component of clinical decision-making \cite{med-reasoning-foundations}. It entails integrating medical knowledge, patient information, and contextual factors to develop accurate diagnoses and effective management plans. Recent advancements in LLMs have drawn increasing attention to their use in medical reasoning \cite{gpt4o, med-palm2, baichuan-m1}. A large body of research has focused on enhancing the medical reasoning capabilities of LLMs through further pre-training with additional medical knowledge or instruction tuning on question-answering datasets, such as MEDITRON \cite{MEDITRON}, Meerkat \cite{meerkat}, and MedAdapter \cite{medadapter}. More recently, reinforcement learning has been applied to improve test-time scaling performance in medical reasoning tasks \cite{huatuogpt-o1, med-rlvr, m1, FineMedLM-o1}.

\subsection{Medical RAG}

Retrieval-Augmented Generation (RAG) plays a crucial role in medical applications due to the knowledge-intensive and high-stakes nature of the domain \cite{medrag}. Recent research efforts focus on refining queries to access more relevant information. Specifically, Med-R$^2$ \cite{med-r2} and RGAR \cite{rgar} enhance retrieval by iteratively modifying medical queries based on generation outcomes. SeRTS \cite{serts} optimizes query generation using Monte Carlo Tree Search with document-relevance feedback. 
Additionally, some work focuses on improving knowledge construction and retrieval-based knowledge usage. \citet{med_textbooks} builds a RAG pipeline with query enhancement and knowledge filtering. MedGraphRAG \cite{medical-graph-rag} constructs a local Knowledge Graph (KG) from medical queries and efficiently retrieves relevant subgraphs. KARE \cite{kare} constructs a multi-source KG by integrating a medical corpus and LLM-generated insights. SPO \cite{omni-rag} investigates source planning and multi-source utilization. However, few studies have explicitly constructed and optimized the reasoning process to improve retrieval. 

\subsection{Reasoning-augmented Retrieval}

Reasoning-intensive tasks present greater challenges to retrieval, primarily because critical knowledge requirements are often concealed within surface information and require reasoning to uncover \cite{bright}. For example, in medical diagnosis tasks, surface information (e.g., age, gender, and examination results) requires further analysis to establish standardized knowledge requirements (e.g., symptoms of hypertension during pregnancy). Although critical, improving retrieval for reasoning-intensive tasks has attracted relatively little research attention. RAG$^2$ \cite{rationale-guided-rag} applies RAG to medical question answering, where queries are augmented with LLM-generated rationales. JudgeRank \cite{judgerank} designs query and document analysis modules to enhance relevance judgment and improve document reranking. Search-R1 \cite{search-r1} learns to generate a series of search queries during step-by-step reasoning with real-time retrieval. RARE \cite{rare} applies MCTS to generate queries. To the best of our knowledge, we are the first to jointly optimize reasoning-augmented retrieval and retrieval-augmented reasoning, and to perform a single retrieval instead of multiple time-consuming retrieval steps.
\section{Conclusion}

In this work, we propose RAR$^2$, a novel joint learning framework designed to enhance LLM reasoning capabilities within the RAG framework, specifically targeting complex medical questions. Unlike existing RAG approaches, which rely primarily on direct query refinement, RAR$^2$ explicitly constructs and optimizes a thought-based reasoning process to uncover implicit knowledge requirements. We introduce a mixed preference dataset comprising thought pairs and answer pairs and leverage DPO for joint training. Extensive experiments on multiple biomedical question answering datasets demonstrate that RAR$^2$ outperforms state-of-the-art RAG baselines, both with and without fine-tuning. Moreover, our analysis of test-time scaling strategies validates the scalability and robustness of RAR$^2$, highlighting its potential to significantly improve medical reasoning and decision-making in clinical scenarios.

\section*{Limitations}

Our work has several limitations. First, the medical corpus we use does not incorporate knowledge graphs. As an important source of structured medical knowledge, knowledge graphs could help address the lack of structure in our current corpus by introducing explicit entity relationships and semantic hierarchies, thereby improving the retrieval of clinically relevant information and supporting more accurate reasoning. Second, our optimization of thought process and answer generation does not involve step-level supervision. Step-level DPO has shown promising results in various domains and represents a worthwhile direction for future exploration. Lastly, our method does not aim to improve the retrieval model itself. Enhancing the retrieval model’s reasoning and understanding capabilities could further unlock the potential of reasoning-augmented retrieval.

\section*{Ethic Statement}

Our proposed framework aims to improve retrieval-augmented medical reasoning and focuses on medical question answering. All datasets used for training and evaluation have been anonymized, and there is no risk of privacy exposure. However, when using LLMs for medical tasks, it is important to be aware that LLMs are prone to hallucinations, and their suggestions should not be considered definitive diagnostic conclusions. Medical advice generated by LLMs must be reviewed by qualified healthcare professionals. Therefore, we do not recommend the direct use of LLMs for medical diagnosis or decision-making at this stage. 
Furthermore, the scientific artifacts that
we used are freely available for research, including Transformers, PyTorch and other GitHub codes. And this paper’s use of these artifacts is consistent with their intended use.

\bibliography{custom}

\appendix
\section{Appendix}

\subsection{Package Details}
We used the following Python packages and their corresponding versions: transformers 4.44.2, pytorch 2.4.0, and xformers 0.0.27.post2.

\subsection{Baseline Implementations}
We adopt the officially released model checkpoints for baseline methods: m1-7B-23K\footnote{\url{https://huggingface.co/UCSC-VLAA/m1-7B-23K}}, HuatuoGPT-o1-7B\footnote{\url{https://huggingface.co/FreedomIntelligence/HuatuoGPT-o1-7B}}, MedRAG (i-MedRAG)\footnote{\url{https://github.com/Teddy-XiongGZ/MedRAG}}, Self-BioRAG\footnote{\url{https://github.com/dmis-lab/self-biorag}}.

\subsection{Details of Evaluation Datasets}
We present the statistics of evaluation datasets in the Table. 

\begin{table}[t!]
    \centering
    \resizebox{0.6\linewidth}{!}{
    \begin{tabular}{@{}lc@{}}
    \toprule
    Datasets   & Number  \\ \midrule
    MedQA      & 1273    \\ 
    MedMCQA    & 4183    \\ 
    MMLU-Med   & 1089    \\ 
    GPQA       & 105     \\ 
    Medbullets & 308     \\
    MedXpertQA & 2450    \\ \bottomrule
    \end{tabular}
    }
    \caption{The number of samples for each evaluation dataset}
    \label{eval_dataset}
    \end{table}

\end{document}